\RequirePackage{fix-cm}
\documentclass[journal]{IEEEtran}

\ifCLASSINFOpdf
  \usepackage[pdftex]{graphicx}
\else
\fi
\usepackage{url}
\usepackage[pdftex,
CJKbookmarks=true, bookmarksnumbered=true, bookmarksopen=true,
colorlinks=true, 
pdfborder=001, 
citecolor=blue, linkcolor=blue, anchorcolor=red, urlcolor=black
]{hyperref}

\usepackage{amssymb}
\usepackage{algorithm}
\usepackage{algpseudocode}
\usepackage[cmex10]{amsmath}

\usepackage{array}

\usepackage{bbding}
\usepackage[cmex10]{amsmath}
\usepackage{url}
\usepackage{color}
\usepackage{hyphenat}
\usepackage{multirow}

\usepackage[tight,footnotesize]{subfigure}

\begin{document}
%
\title{Training-Free Synthesized Face Sketch Recognition Using Image Quality Assessment Metrics}
%
%
%

\author{Nannan~Wang,~\IEEEmembership{Member,~IEEE,}
        Jie~Li,
        Leiyu~Sun,
        Bin~Song,
        and Xinbo~Gao,~\IEEEmembership{Senior Member,~IEEE}
%
}

\maketitle

\begin{abstract}
Face sketch synthesis has wide applications ranging from digital entertainments to law enforcements. Objective image quality assessment scores and face recognition accuracy are two mainly used tools to evaluate the synthesis performance. In this paper, we proposed a synthesized face sketch recognition framework based on full-reference image quality assessment metrics. Synthesized sketches generated from four state-of-the-art methods are utilized to test the performance of the proposed recognition framework. For the image quality assessment metrics, we employed the classical structured similarity index metric and other three prevalent metrics: visual information fidelity, feature similarity index metric and gradient magnitude similarity deviation. Extensive experiments compared with baseline methods illustrate the effectiveness of the proposed synthesized face sketch recognition framework. Data and implementation code in this paper are available online at \url{www.ihitworld.com/WNN/IQA_Sketch.zip}.
\end{abstract}


\IEEEpeerreviewmaketitle

\section{Introduction}
\label{sec:Introduction}
\IEEEPARstart{F}{ace} sketch synthesis is demanded in many aspects in real-world applications, \textit{e.g.} digital entertainment and law enforcement \cite{IJCVSurvey}. For digital entertainment, many people would like to take sketch portrait generated from a photo as the profile for their social network accounts. And also, recently face sketch synthesis has been applied to 3D Chocolate printer for the purpose of printing a black and white sketch as the guidance for printing.

\begin{figure}
\centering
\includegraphics[width=1\columnwidth]{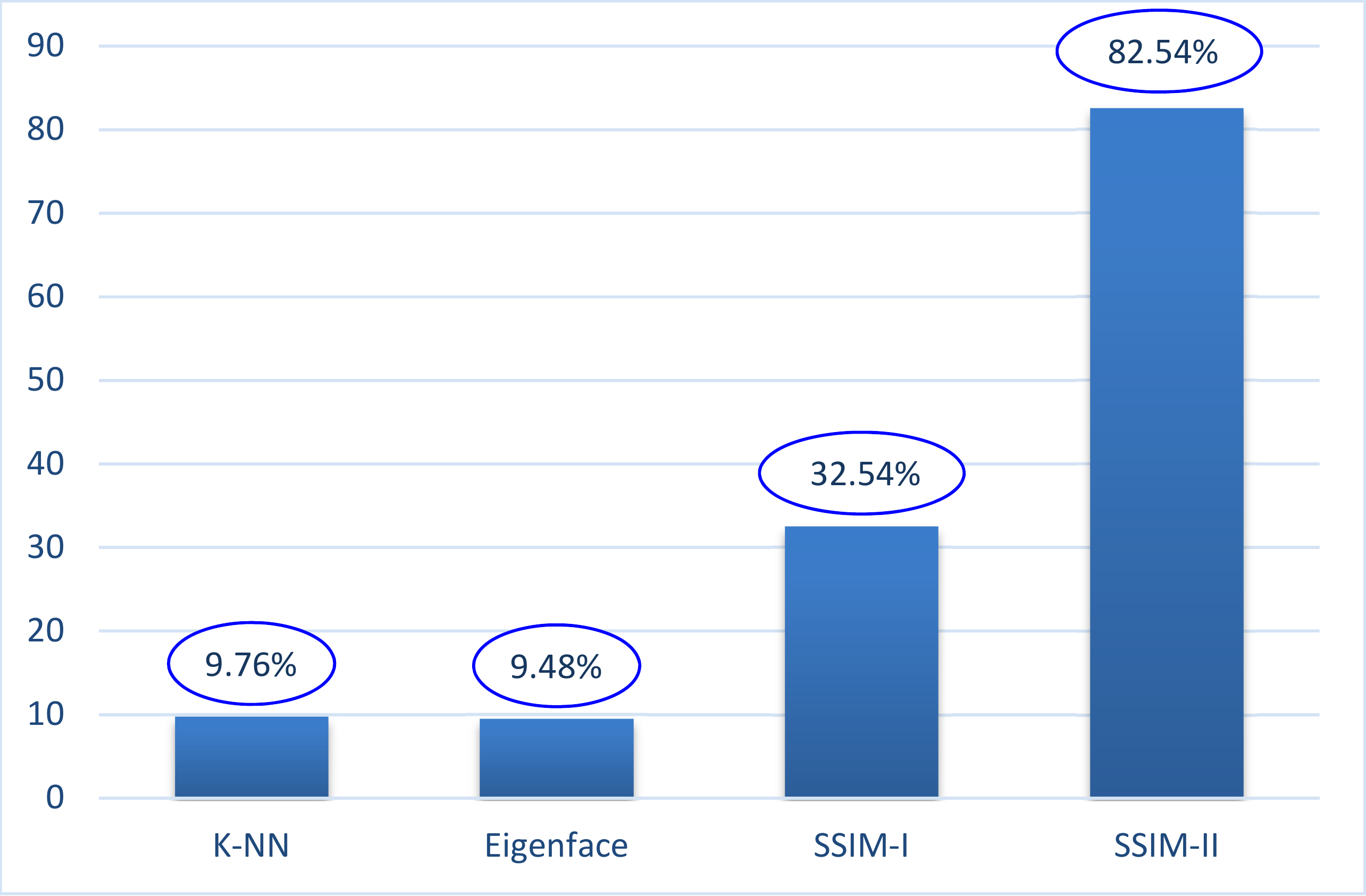}
\caption{Face recognition accuracy by direct matching a sketch to the mug shot (K-NN, Eigenface, and SSIM-I) vs. by using the proposed framework (SSIM-II).}
\label{fig:1}
\end{figure}

Another application in law enforcement is inspired by the fact that a photo of the suspect is not always available due to their deliberately avoidance. A sketch drawn by the artist according to the descriptions of the eyewitness or clues from surveillance videos could be a substitute. However, due to the great discrepancy in texture and their imaging modes, directly matching the sketch to the mug shot performs poorly. Fig. \ref{fig:1} gives the comparison between directly matching and matching using the proposed framework. "$K$-NN" represents the $K$ nearest neighbors and it is set to 1 for this face recognition experiment. Eigenface \cite{Eigenface} refers to project both the mug shot and the test sketch into a subspace and then perform $1$-NN. "SSIM" is the abbreviation of structured similarity index \cite{SSIM}. "SSIM-I" is to perform $1$-NN among SSIM scores of the sketch and all photos in mug shot database. "SSIM-II" is an implementation within our proposed framework taking SSIM as the image quality assessment (IQA) metric. It can be seen that directly matching a sketch to the mug shot database fails as shown in Fig. \ref{fig:1} while the proposed framework could achieve much better performance benefited from the face sketch synthesis procedure. The face sketch synthesis procedure is to decrease the discrepancy between the sketch and mug shot photos. In real-world applications, we can transforming all photos in the mug shot database into sketches and then the probe sketch drawn by the artist can be matched on the synthesized sketch database.

\begin{figure*}
\centering
\includegraphics[width=1.9\columnwidth]{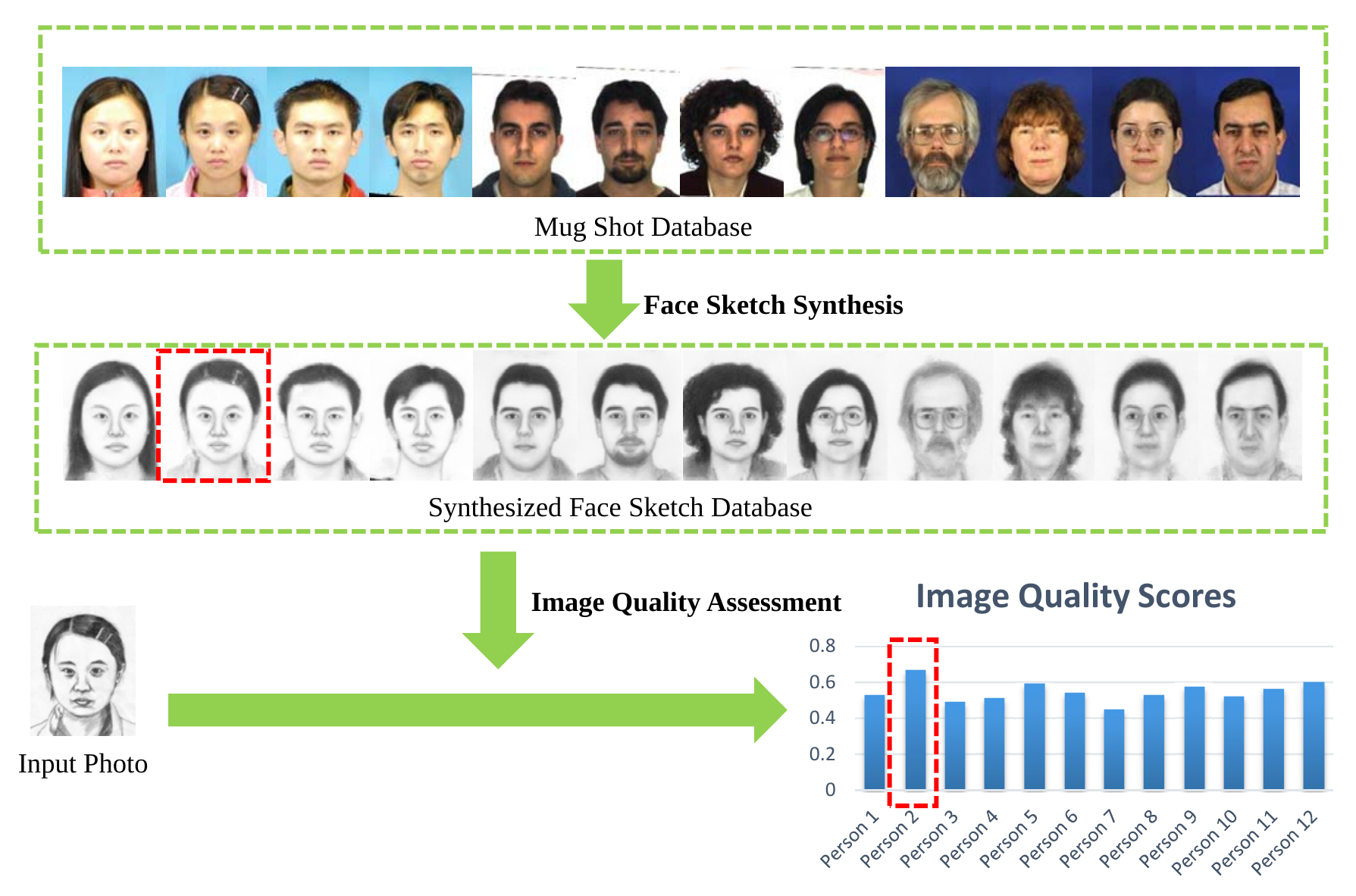}
\caption{Graphical outline of the proposed framework.}
\label{fig:2}
\end{figure*}

There are many softwares which could generate a sketch by feeding into a photo. However, it has been shown that these image-based methods cannot  mimic the sketch style and their results are more like photos \cite{SSD}. Starting from the work of Tang and Wang \cite{ICIP2002}, exemplar-based face sketch synthesis has attracted growing attentions. This category of methods could learn the drawing style such as shadow and texture more vividly. Except the starting work \cite{ICIP2002} which employs principal component analysis to compute the reconstruction coefficient in a holistic manner, existing methods work on patch-level. Given a test photo, it is firstly divided into some patches in the same way as the training sketch-photo pairs. Then for each test patch, $K$ or some number of nearest photo patch neighbors are searched from the training photos. The target sketch patch is synthesized by linearly combining sketch patch candidates corresponding to the selected photo patch neighbors. Finally these target sketch patches are assembled into a whole image by averaging or quilting \cite{quilting} the overlapping region.

These methods could be classified into two groups: methods synthesizing each sketch patch independently and methods taking neighboring constraint into consideration. The representative algorithms in the former group include the locally linear embedding (LLE) method \cite{LLE}, sparse neighbor selection method \cite{SNS}, the spatial sketch denosing (SSD) method \cite{SSD} and so on. The laster group mainly refers to probabilistic graphical model based methods, \textit{e.g.} the Markov random filed (MRF) method \cite{MRF} and the Markov weight filed (MWF) method \cite{MWF}.

IQA metrics and face recognition metrics are commonly used to evaluate the performance of face sketch synthesis methods \cite{IJCVSurvey}. SSIM \cite{SSIM} and Eigenface \cite{Eigenface} are the respective representative method for these two evaluations. IQA metric provides a measure to evaluate the quality of synthesized sketch in the manner of image distortions. Face recognition accuracy can be seemed as an indirect way to assess the performance of face sketch synthesis. The assumption behind this is that higher face recognition accuracy it reaches, better performance the face sketch synthesis method achieves. In this paper, we proposed a new evaluation framework by embedding the IQA metrics into the face recognition framework. The proposed framework could simultaneously evaluate the quality of synthesized sketches and conduct the face recognition application. In addition, since it is expensive to collect large scale of sketches for learning the classifier, the proposed method is training-free. The details of the proposed framework will be given in next section.

\section{IQA Metrics for Synthesized Face Sketch Recognition}

The graphical outline for the proposed framework is shown in Fig. \ref{fig:2}. Firstly, all mug shot photos are transformed into sketches by face sketch synthesis algorithms. Secondly, given a probe sketch, it is taken as the reference image and each synthesized sketch in the gallery is taken as the distorted image. IQA scores are obtained by full-reference IQA metrics, \textit{e.g.} SSIM. Finally, the synthesized sketch which obtain the largest IQA score is identified as the suspect in a $1$-NN manner.

In this paper, we employ four state-of-the-art face sketch synthesis methods (two independent methods and two Bayesian methods) to transform mug shot photos into sketches: the LLE method \cite{LLE}, the SSD method \cite{SSD}, the MRF method \cite{MRF} and the MWF method \cite{MWF}. Results of SSD and MWF are generated form the source code provided by authors. The results of LLE is generated from our implementation and the MRF source code are download from the website: \url{http://www.cs.cityu.edu.hk/~yibisong/eccv14/index.html}.

\begin{figure*}
\centering
\includegraphics[width=1.9\columnwidth]{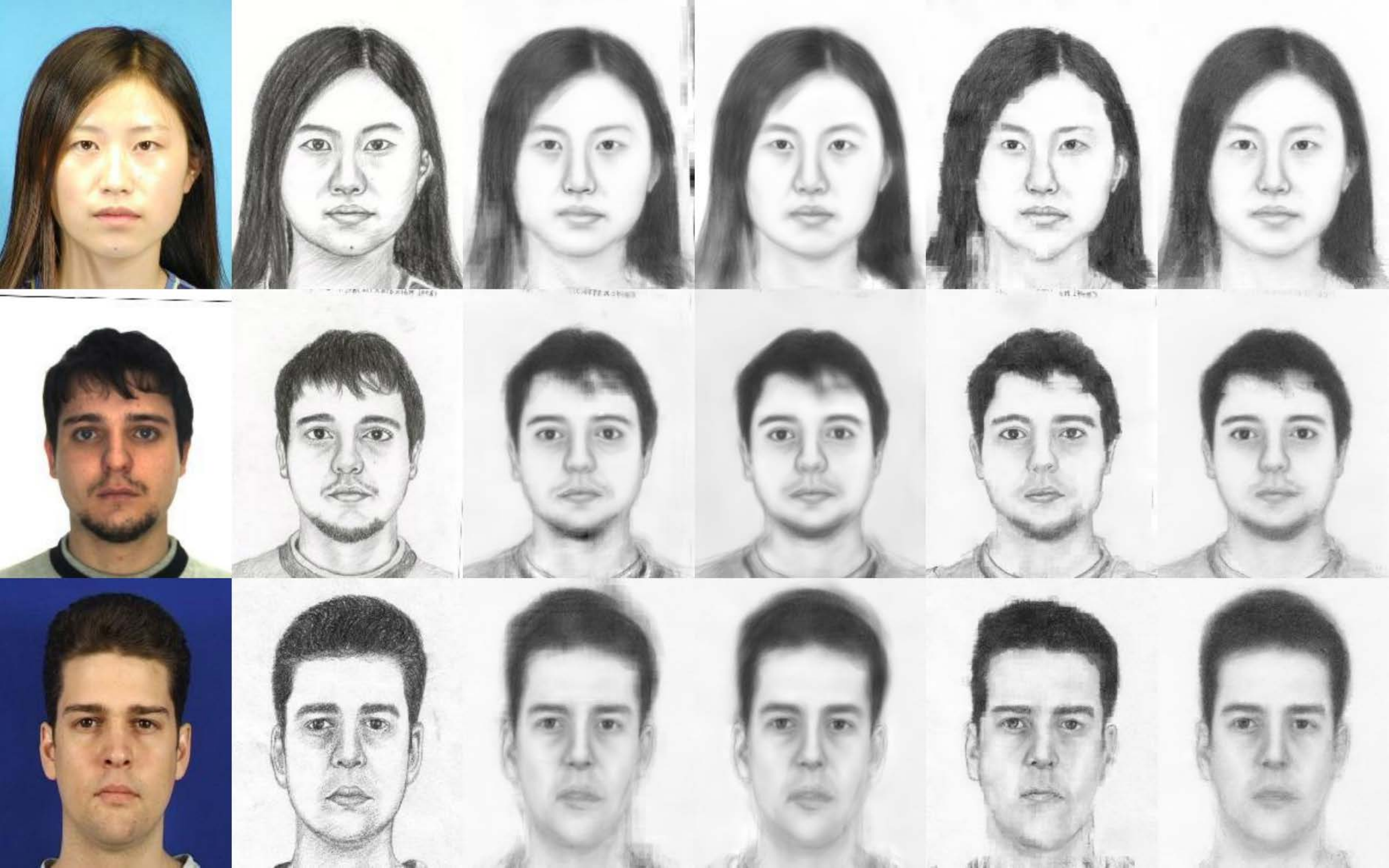}
\caption{Synthesized face sketch examples by four different methods on three datasets. The first column is the input photo and the second column is the corresponding sketch drawn by the artist. The third to the last column are the results of LLE, SSD, MRF and MWF. These three face photos are from the CUHK Studen, AR, and XM2VTS dataset respectively.}
\label{fig:3}
\end{figure*}

Face sketches are from the Chinese University of Hong Kong (CUHK) face sketch database (CUFS). It is composed of three sub-datasets: the CUHK student dataset \cite{ICIP2002}, the AR dataset \cite{AR} and the XM2VTS dataset \cite{XM2VTS}. Some synthesized examples on these three datasets are shown in Fig. \ref{fig:3}. The complete data and evaluation source codes can be downloaded from the website: \url{www.ihitworld.com/WNN/IQA_Sketch.zip}.

\begin{figure*}
\centering
\subfigure[]{
\label{fig:subfig1}
\includegraphics[width=0.47\columnwidth]{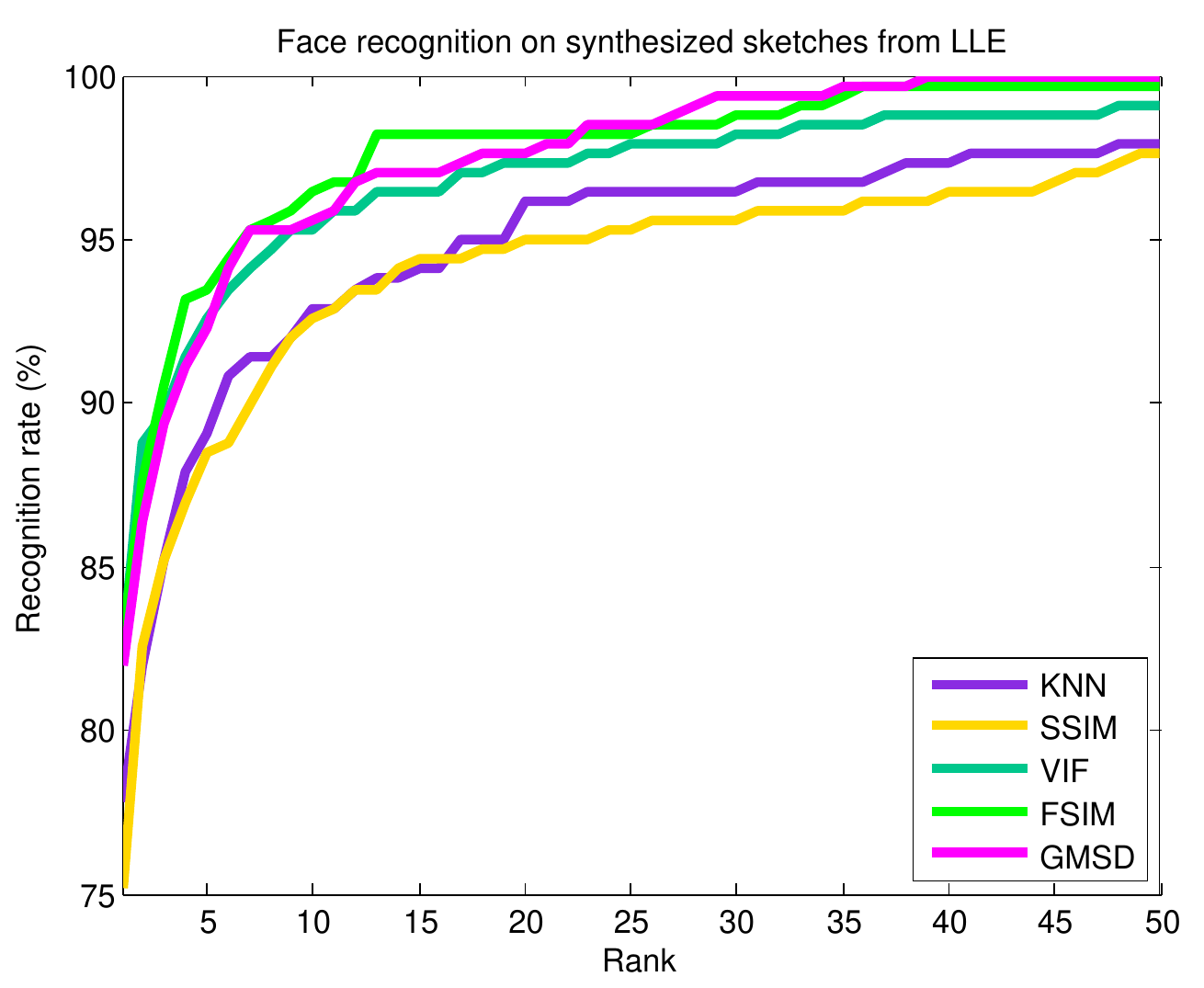}
}
\subfigure[]{
\label{fig:subfig2}
\includegraphics[width=0.47\columnwidth]{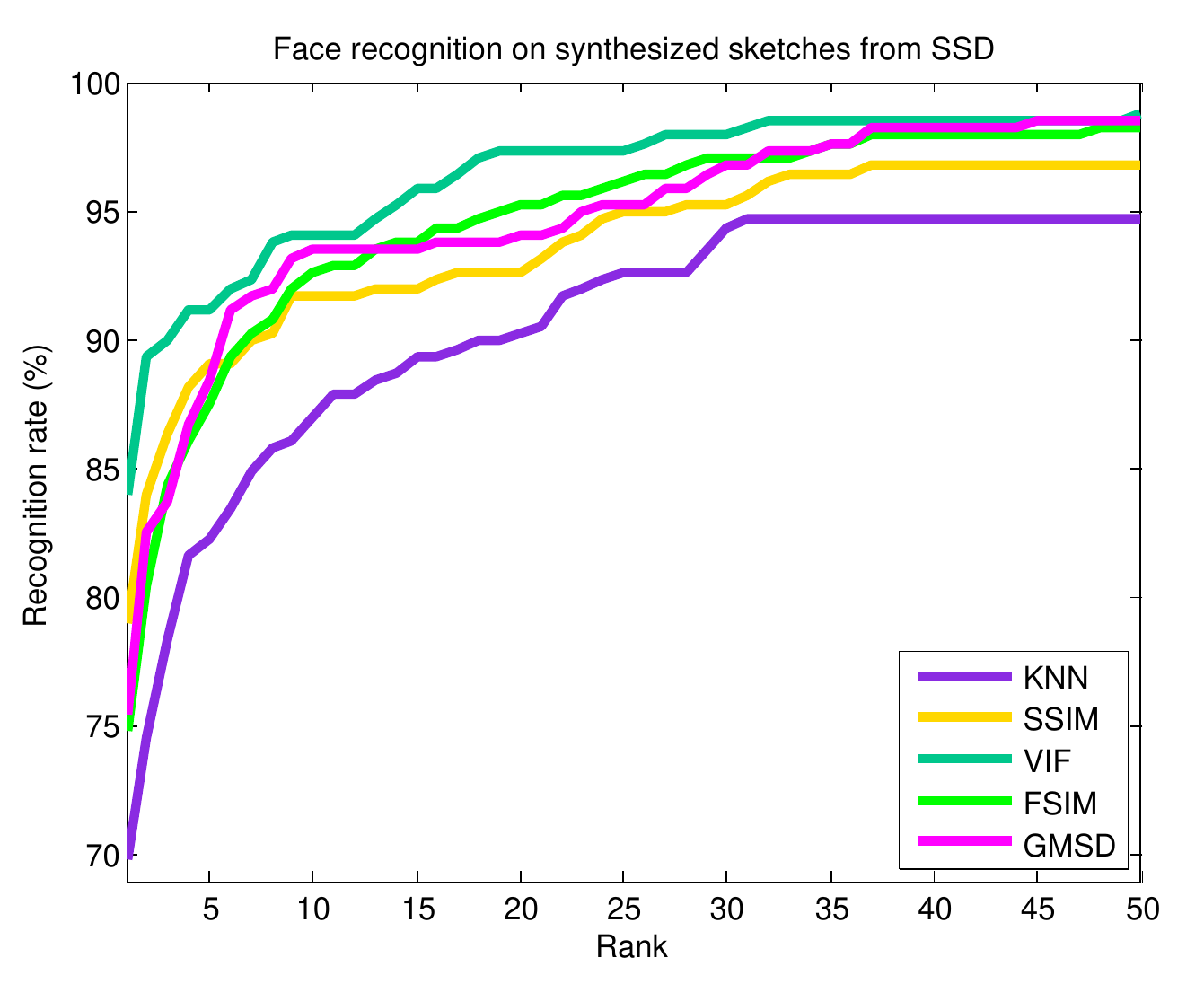}
}
\subfigure[]{
\label{fig:subfig3}
\includegraphics[width=0.47\columnwidth]{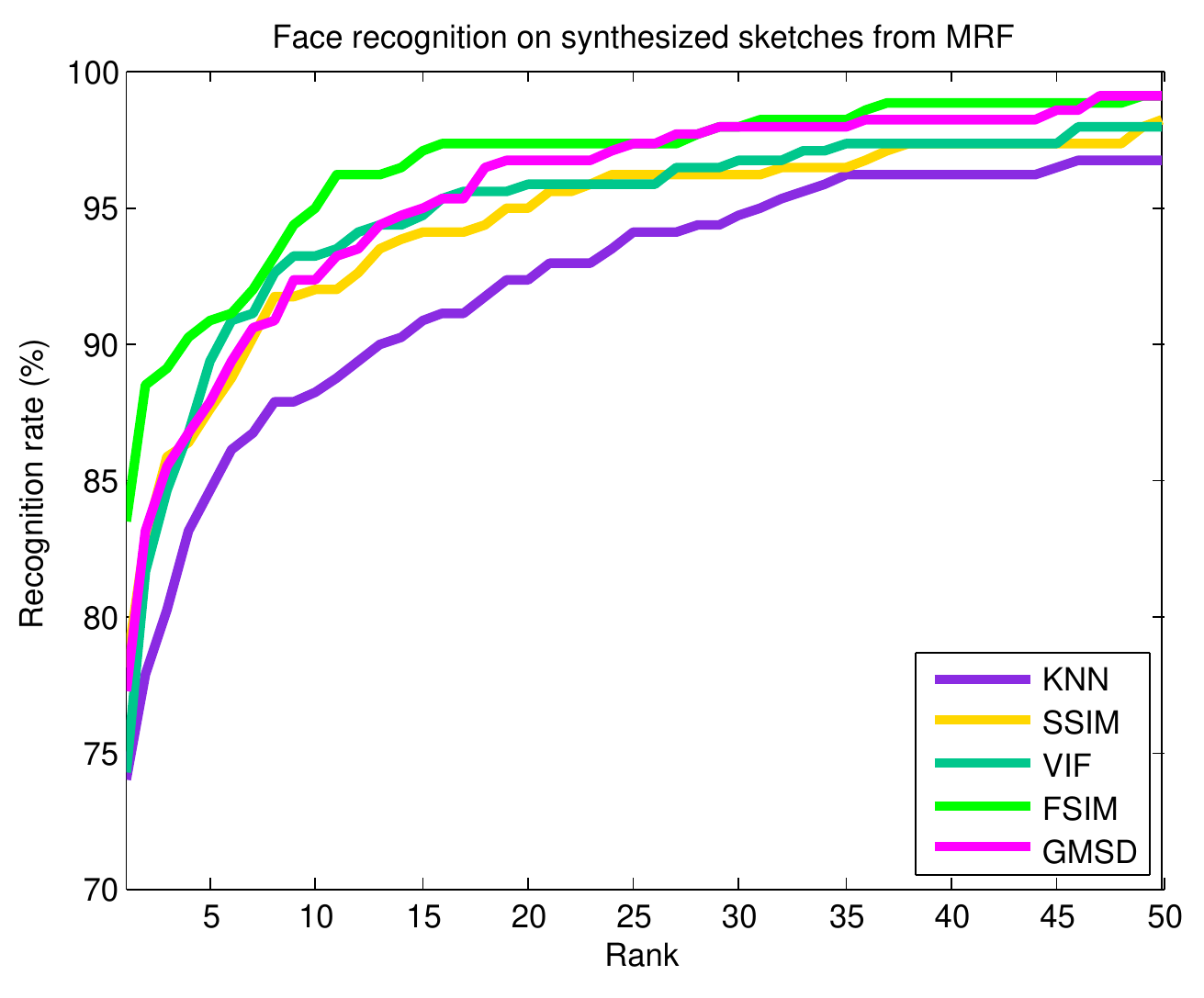}
}
\subfigure[]{
\label{fig:subfig4}
\includegraphics[width=0.47\columnwidth]{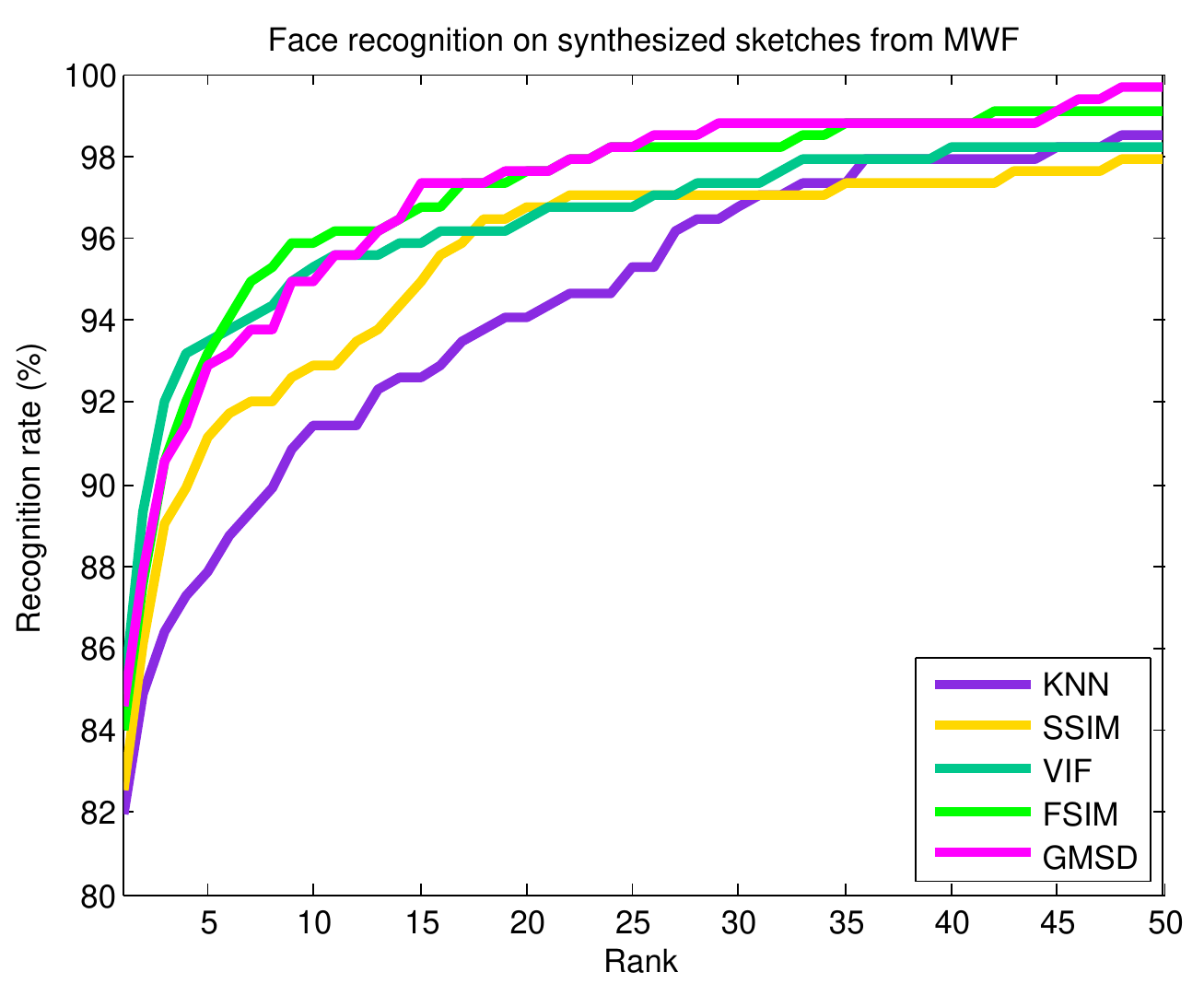}
}
\caption{Face recognition using different image quality assessment metrics and $K$-NN on sketches generated by (a) LLE, (b) SSD, (c) MRF, (d)MWF}
\label{fig:4}
\end{figure*}

To conduct the full-reference IQA, besides the classical method SSIM \cite{SSIM}, we also employ three state-of-the-art methods:  visual information fidelity (VIF) \cite{VIF}, feature similarity index metric (FSIM) \cite{FSIM} and gradient magnitude similarity deviation (GMSD) \cite{GMSD}.

$K$-NN is taken as the baseline recognition method since it is also training-free. Fig. \ref{fig:4} presents the cumulative face recognition accuracy (rank 1 to rank 50) between $K$-NN and IQA-based method on sketches generated by aforementioned four methods. From Fig. \ref{fig:4}, it can be seen that generally IQA-based methods achieve better performance than the $K$-NN method. Except the SSIM-based method, other three IQA metric based face recognition methods outperform $K$-NN a lot. Even the SSIM-based method could obtain comparable or better performance than the $K$-NN strategy.

\begin{table*}
\centering
\caption{Cumulative match accuracies using SSIM as the recognition metric}
\label{tab:table 1}
\begin{tabular}{cccccccccccccccc}
\hline
 Rank  & 1 & 2 & 3 & 4 & 5 & 6 & 7 & 8 & 9 & 10 & 11 & 12 & 13 &14 & 15  \\
\hline
LLE (\%) &  75.15 & 82.54 & 85.21 & 86.98 & 88.46 & 88.76 & 89.94 & 91.12 & 92.01 & 92.60 & \textbf{92.90} & \textbf{93.49} & 93.49 & 94.08 & 94.38  \\
SSD (\%) &  78.99 & 84.02 & 86.39 & 88.17 & 89.05 & 89.05 & 89.94 & 90.24 & 91.72 & 91.72 & 91.72 & 91.72 & 92.01 & 92.01 & 92.01  \\
MRF (\%) &  78.11 & 82.84 & 85.80 & 86.39 & 87.57 & 88.76 & 90.24 & 91.72 & 91.72 & 92.01 & 92.01 & 92.60 & 93.49 & 93.79 & 94.08  \\
MWF (\%) &  \textbf{82.54} & \textbf{86.09} & \textbf{89.05} & \textbf{89.94} & \textbf{91.12} & \textbf{91.72} & \textbf{92.01} & \textbf{92.01} & \textbf{92.60} & \textbf{92.90} & \textbf{92.90} & \textbf{93.49} & \textbf{93.79} & \textbf{94.38} & \textbf{94.97}   \\
\hline
\end{tabular}
\end{table*}

\begin{table*}
\centering
\caption{Cumulative match accuracies using VIF as the recognition metric}
\label{tab:table 2}
\begin{tabular}{cccccccccccccccc}
\hline
 Rank  & 1 & 2 & 3 & 4 & 5 & 6 & 7 & 8 & 9 & 10 & 11 & 12 & 13 &14 & 15  \\
\hline
LLE (\%) &  82.54 & 88.76 & 89.64 & 91.42 & 92.60 & 93.49 & \textbf{94.08} & \textbf{94.67} & \textbf{95.27} & \textbf{95.27} & \textbf{95.86} & \textbf{95.86} & \textbf{96.45} & \textbf{96.45} & \textbf{96.45}  \\
SSD (\%) &  84.02 & 89.35 & 89.94 & 91.12 & 91.12 & 92.01 & 92.31 & 93.79 & 94.08 & 94.08 & 94.08 & 94.08 & 94.67 & 95.27 & 95.86  \\
MRF (\%) &  74.26 & 81.66 & 84.62 & 86.69 & 89.35 & 90.83 & 91.12 & 92.60 & 93.20 & 93.20 & 93.49 & 94.08 & 94.38 & 94.38 & 94.67  \\
MWF (\%) &  \textbf{84.91} & \textbf{89.35} & \textbf{92.01} & \textbf{93.20} & \textbf{93.49} & \textbf{93.79} & \textbf{94.08} & 94.38 & 94.97 & \textbf{95.27} & 95.56 & 95.56 & 95.56 & 95.86 & 95.86   \\
\hline
\end{tabular}
\end{table*}

\begin{table*}
\centering
\caption{Cumulative match accuracies using FSIM as the recognition metric}
\label{tab:table 3}
\begin{tabular}{cccccccccccccccc}
\hline
 Rank  & 1 & 2 & 3 & 4 & 5 & 6 & 7 & 8 & 9 & 10 & 11 & 12 & 13 &14 & 15  \\
\hline
LLE (\%) &  82.84 & 87.57 & 90.53 & \textbf{93.20} & \textbf{93.49} & \textbf{94.38} & \textbf{95.27} & \textbf{95.56} & \textbf{95.86} & \textbf{96.45} & \textbf{96.75} & \textbf{96.75} & \textbf{98.22} & \textbf{98.22} & \textbf{98.22}  \\
SSD (\%) &  74.85 & 80.47 & 84.32 & 86.09 & 87.57 & 89.35 & 90.24 & 90.83 & 92.01 & 92.60 & 92.90 & 92.90 & 93.49 & 93.79 & 93.79  \\
MRF (\%) &  83.43 & 88.46 & 89.05 & 90.24 & 90.83 & 91.12 & 92.01 & 93.20 & 94.38 & 94.97 & 96.15 & 96.15 & 96.15 & 96.45 & 97.04  \\
MWF (\%) &  \textbf{84.02} & \textbf{87.57} & \textbf{90.53} & 92.01 & 93.20 & 94.08 & 94.97 & 95.27 & \textbf{95.86} & 95.86 & 96.15 & 96.15 & 96.15 & 96.45 & 96.75   \\
\hline
\end{tabular}
\end{table*}

\begin{table*}
\centering
\caption{Cumulative match accuracies using GMSD as the recognition metric}
\label{tab:table 4}
\begin{tabular}{cccccccccccccccc}
\hline
 Rank  & 1 & 2 & 3 & 4 & 5 & 6 & 7 & 8 & 9 & 10 & 11 & 12 & 13 &14 & 15  \\
\hline
LLE (\%) &  81.95 & 86.39 & 89.35 & 91.12 & 92.31 & \textbf{94.08} & \textbf{95.27} & \textbf{95.27} & \textbf{95.27} & \textbf{95.56} & \textbf{95.86} & \textbf{96.75} & \textbf{97.04} & \textbf{97.04} & 97.04  \\
SSD (\%) &  75.44 & 82.54 & 83.73 & 86.69 & 88.46 & 91.12 & 91.72 & 92.01 & 93.20 & 93.49 & 93.49 & 93.49 & 93.49 & 93.49 & 93.49  \\
MRF (\%) &  77.22 & 83.14 & 85.50 & 86.69 & 87.87 & 89.35 & 90.53 & 90.83 & 92.31 & 92.31 & 93.20 & 93.49 & 94.38 & 94.67 & 94.97  \\
MWF (\%) &  \textbf{84.62} & \textbf{87.87} & \textbf{90.53} & \textbf{91.42} & \textbf{92.90} & 93.20 & 93.79 & 93.79 & 94.97 & 94.97 & 95.56 & 95.56 & 96.15 & 96.45 & \textbf{97.34}   \\
\hline
\end{tabular}
\end{table*}

Besides the face recognition application, the proposed IQA-based training-free framework could also utilized to compare the performance of different face sketch synthesis methods. Table \ref{tab:table 1}-\ref{tab:table 4} list the cumulative face recognition accuracy (due to the space limit, here we only show rank-1 to rank-15) by different IQA metrics. Generally, the MWF method \cite{MWF} could achieve better performance among first several ranks and the LLE method \cite{LLE} could achieve better performance among last several ranks. The best rank-1 accuracy (84.91\%) is achieved by the VIF metric based recognition method on synthesized sketches generated by the MWF method. The best rank-50 accuracy is 98.22\% achieved by the FSIM metric based recognition strategy on sketches generated by the LLE method.

To further illustrate the effectiveness of the proposed method, we compared the proposed framework with a training-based face recognition method: the Eigenface method \cite{Eigenface}. We randomly select 150 synthesized face sketches and their corresponding sketch drawn by the artist as the training data. The rest 188 synthesized face sketches as the gallery set and the 188 corresponding sketch drawn by the artist is taken as the probe image. This process is repeated 100 times and the average accuracy is reported in this paper. Table \ref{tab:table 5} compares the Eigenface method with four IQA metric based recognition methods. It can be seen that the proposed IQA metric based recognition methods could achieve better performance than the Eigenface method. This further validates the effectiveness of the proposed framework.

\begin{table}
\centering
\caption{The best accuracy of the Eigenface method vs. accuracies of the proposed framework}
\label{tab:table 5}
\begin{tabular}{cccccc}
\hline
 Method & SSIM & VIF & FSIM & GMSD & Eigenface  \\
\hline
LLE (\%) &  75.15 & 82.54 & 82.84 & 81.95 & 77.80\\
SSD (\%) &  78.99 & 84.02 & 74.85 & 75.44 & 71.34\\
MRF (\%) &  78.11 & 74.26 & 83.43 & 77.22 & 71.33\\
MWF (\%) &  82.54 & 84.91 & 84.02 & 84.62 & 80.21\\
\hline
\end{tabular}
\end{table}

Aforementioned experiments demonstrate that the proposed training-free recognition framework based on IQA metrics could achieve better performance than commonly used $K$-NN method, another training-free method. In addition, the proposed framework even outperforms the classical training based method such as Eigenface \cite{Eigenface}. Almost all existing IQA metrics can be embedded into the proposed framework and customized IQA metric for synthesized face sketch may be even better. Actually, the proposed framework could be generalized to other heterogeneous face image transformation applications \cite{HIT} such as the transformation between visible face images and near infrared images.

\section{Conclusion}
In this paper, we proposed a training-free synthesized face sketch recognition framework based on image quality assessment metrics. Four full-reference image quality assessment metrics are employed to implement the framework: one classical method (SSIM) and three prevalent methods (VIF, FSIM, and GMSD). Four state-of-the-art face sketch synthesis methods are utilized to transform the photos in the mug shot database into sketches. Experimental results illustrate that the proposed framework perform better than the baseline approach: $K$-NN. We also compared the proposed IQA metric based recognition framework with the training based method (Eigenface) and the superior performance validate the effectiveness of the proposed framework. In addition, the proposed face recognition framework can also be employed as the evaluation metric to assess the performance of different face sketch synthesis methods. In the future, we would apply the proposed framework to evaluate the performance of much more heterogeneous face image transformation applications.

\ifCLASSOPTIONcaptionsoff
  \newpage
\fi

\bibliographystyle{IEEEtran}
\end{document}